\documentclass[journal,twocolumn,10pt]{IEEEtran}
\usepackage{multirow}
\usepackage{amsmath}
\usepackage{setspace}
\usepackage{amsthm}
\usepackage{amscd}
\usepackage{hhline}
\usepackage[dvips]{graphicx} 
\usepackage{amssymb}
\usepackage{color}
\usepackage[table]{xcolor}

\usepackage{multicol}

\begin{document}


	\title{ 
		\huge Biometrics in the Time of Pandemic: 40\% Masked Face Recognition Degradation
		can be Reduced to 2\%
		\thanks{}
	}

	\author{L. Queiroz, K. Lai, S. Yanushkevich, and V. Shmerko 
		\thanks{
			Authors are with Biometric Technologies Laboratory, Department of Electrical \& Software
			Engineering, University of
			Calgary, Canada, Web: http://www.ucalgary.ca/btlab. E-mail:
			\{leonardo.queiro,kelai,syanshk,vshmerko\}@ucalgary.ca. 
		}
	}
	
	

	\maketitle
	

	\begin{abstract}
		In this study of the face recognition on masked versus unmasked faces generated using Flickr-Faces-HQ and SpeakingFaces datasets, we report 36.78\% degradation of recognition performance caused by the mask-wearing at the time of pandemics, in particular, in border checkpoint scenarios. We have achieved better performance and reduced the degradation to 1.79\% using advanced deep learning approaches in the cross-spectral domain.

		
		
	\end{abstract}

	\textbf{Keywords:} Face biometrics, periocular biometrics, face recognition, thermal imaging, cross-spectrum, face mask, border control, deep machine learning, generative adversarial network.
	
	\section{Introduction}\label{sec:}
	
\begin{figure}[!h]
		\begin{center}
			\includegraphics[width=0.45\textwidth]
			{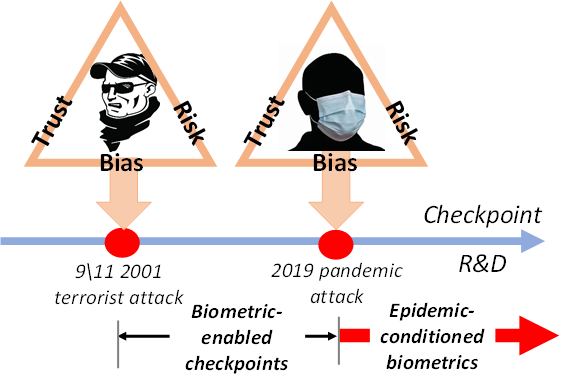}
			\caption{Two main events drastically impact the R\&D of the security checkpoints: the 2001 9/11 event and the 2019 COVID-19 pandemic. The 9/11 terrorist attack has accelerated the R\&D of the biometric-enabled checkpoints. The COVID-19 pandemic attack has triggered the R\&D of the epidemic-conditioned biometrics such as periocular facial biometrics for the face mark wearing travelers.
			}\label{fig:RTB_Face}
		\end{center}
	\end{figure}

	\IEEEPARstart{B}{}iometric-enabled security checkpoints deployed in mass-transit hubs such as airports and seaports are the frontiers of national and international security \cite{[IATA-ABC-Guide-2015],[TSA-2006],[TSA-2013]}. Two events had a drastic impact on the Research \& Development (R\&D) of security checkpoints: the 9/11 2001 terrorist attack and the 2019 COVID-19 pandemic (Fig. \ref{fig:RTB_Face}). 
	
	Post 9/11 R\&D period focused on increasing security, by introducing biometric-enabled ID and trusted traveler service \cite{[Avoine-2016],[ICAO-e-passport-2015]}, self-service kiosks (gates) \cite{[ISO/IEC-29195],[Nuppeney-EasyPASS-2014]},  biometric-enabled watchlist screening \cite{[Lai-2017],[Yanush-2019],[Yanush-Imperson-2015]}, identity de-duplication detection \cite{[Sudhish-2016]}, authentication machines \cite{[Eastwood-IEEE-J-2015],[Labati-2016]}, implementation and deployment guides and regulations \cite{[Nuppeney-EasyPASS-2014],[IATA-ABC-Guide-2015]}, roadmapping \cite{[IATA-Checkpoint-future]}, human right protection \cite{[Clavell-2017]}, as well as harmonization of advanced management techniques in order to increase a checkpoint performance \cite{[Albert-2021]}. The key performance measures remained the travelers' satisfaction with the service such as time of authentication, public acceptance of the biometric traits (face, iris, fingerprints), waiting times, privacy issues, and addressing the language barriers. 
	
	The full spectrum of post-pandemic challenges is yet to be specified. However, the current technological and societal state of security checkpoints during the pandemic has been characterized by 1) degradation of biometric-enabled technologies such as facial recognition because of mask wearing, and 2) emergence of the ICAO counter-epidemiological initiatives such as immunity passport \cite{[IATA-Travel-pass-2021],[IATA-Travel-pass-Future-travel-2021]}. Performance measures such as risk, trust, and bias are of particular interest, for example, risk of mis-identification when using immunity passport, masked face recognition bias due to the unsatisfactory amount of training data, as well as traveler trustworthiness to accomplished technologies (e.g. tracing), and bias of decision-making (e.g. reliability of infection testing). 
	
	The counter-epidemic response by IATA (International Air Transport Association) prompted the deployment of the travel-pass screening mechanism \cite{[IATA-Travel-pass-2021],[IATA-Travel-pass-Future-travel-2021]} as a step of critical importance in transferring from a centralized platform to the trusted distributed platform. This is only a part of a technological breakthrough that addresses counter epidemic measures. The next step is combining the biometric technology, e.g. transferring biometric-enabled technologies onto distributed platforms. 
	
	Recent work on epidemic-conditioned biometrics focused on periocular biometrics (periocular is the face area around the eyes) \cite{[Narang-2020],[Alonso-Fernandez-2020],[Ngan-2020]}, mask detection \cite{[Qin-2020]}, and adjusting e-interviewers to the acoustic effects of mask wearing \cite{[Nguyen-2021]}. We also refer to \cite{[Budd-2020]} for legal, ethical, and privacy concerns of the counter-epidemic checkpoint extension. In particular, the counter-epidemic checkpoint will need to be 'proofed' against invasions of privacy. The data sharing will need to satisfy the requirements for an independent audit, to ensure data are not used for purposes outside of the pandemic. There is also a concern that the emergency checkpoint mode will set a precedent and may remain beyond the end of the state of emergency.
	
	The COVID-19 pandemic has prompted the expansion and accelerated development of \emph{epidemic-conditioned} biometrics. This means that biometrics trait recognition algorithms are under constraints caused by the counter-epidemiological measures, e.g. masks, shields, gloves, and other personal protective equipment. These measures also impede the performance of biometric-enabled tools and systems. The following biometric traits are epidemic-conditioned:
	\begin{itemize}
		\item [$-$] Face appearance obstructed by a mask, safety glasses, and/or shield'; the face mask prompts the usage of the periocular face region.
		\item [$-$] Fingerprint and palmprint (both contact and touchless) usage is prevented by protective gloves or the need to use sanitizer to the point of impracticality.
		\item [$-$] Iris biometrics may be affected by the safety glasses and shield but not the mask alone.
		\item [$-$] Voice biometrics is impeded by the face mask or shield.
		\item [$-$] Affective state is also conditioned by the face mask; its assessment is limited by the periocular face region.
	\end{itemize}

	\section{Degradation of facial recognition and its recovering}\label{sec:Demonstrative-experiments}

	In this Section, we show experimentally how the checkpoint biometric recognition should be adjusted towards the counter epidemiological requirements. For this, we chose two mandatory post-pandemic checkpoint recognition modes: 1) Face mask detection, including whether it is worn correctly, and 2) authentication of the person wearing the mask. The authentication scenario, in this case, is as follows: given a person, 1) their face is acquired and 2) the face features are extracted, and matching is performed against the data stored in the e-ID. The challenge of the pandemic times is that only the periocular part of the face is available while the lower part is hidden by a mask. 
	
	\subsection{Goals and approach}
	
	According to the ICAO-IATA recommendations, there are three kinds of biometric traits implemented within the e-ID: face, fingerprints, and iris \cite{[ICAO-e-passport-2015]}. In epidemiological scenarios, counter-epidemic measures such as personal protective equipment (masks, shields, glasses, and gloves) impact the availability of these biometric traits. Hence, it is reasonable to consider epidemic-conditioned biometric traits. In this paper, we focus on the mask-affected person's face recognition which is rather replaced by a periocular recognition. Intuitively, the performance of the face recognition system will be degraded if only part of the face is available. Thus, the goals of our experiments are as follows:
	
	\begin{enumerate}
		\item [] \hspace{-8mm}\textbf{Goal I:} Estimate the face recognition degradation given that only the periocular region is available; 
		\item [] \hspace{-8mm}\textbf{Goal II:} Investigate whether compensation for this degradation is possible by using additional data. 
	\end{enumerate}
	
	We investigate the potential of additional data such as face image in infrared band for processing the mask-obstructed part of the face. Our approach is as follows: 
	
	\begin{small}
		\begin{center}
			\begin{parbox}[h]{0.95\linewidth} {
					\vspace{-2mm}
					\begin{center}
						$\underbrace{
							\left\{\hspace{-0.2cm}
							\begin{array}{c}
								\texttt{Periocular} \\
								\texttt{Face} \\
								\texttt{Recognition} \\
							\end{array}\hspace{-0.2cm}
							\right\}				
						}_{\textit{\footnotesize Visual band}}~{\rightarrow}
						\underbrace{\left\{\hspace{-0.2cm}
							\begin{array}{c}
								\texttt{Face} \\
								\texttt{Recognition} \\
							\end{array}\hspace{-0.2cm}
							\right\}
						}_{\textit{ \footnotesize Visual band}}			
						{\leftarrow}~
						\underbrace{
							\left\{\hspace{-0.2cm}
							\begin{array}{c}
								\texttt{'Unmasking' } \\
								\texttt{Lower } \\
								\texttt{Face} \\
							\end{array}\hspace{-0.2cm}
							\right\}
						}_{\textit{ \footnotesize Infrared band}}$\\
					\end{center}
			} \end{parbox}
		\end{center}
	\end{small}
	
	There are several rational arguments for this approach: 
	\begin{enumerate}
		\item [$-$] The cross-sensor periocular biometrics have been studied, in particular, in \cite{[Narang-2020],[Alonso-Fernandez-2020],[Ngan-2020],[Ghosh-2020]};
		\item [$-$] Thermal or Infrared (IR) facial image has been used as a source of health-related indicators such as breathing function and air-breathing temperature \cite{[Murthy-2006]}. 
	\end{enumerate}

\begin{figure*}[!h]
		\begin{center}
			\includegraphics[width=0.9\textwidth]{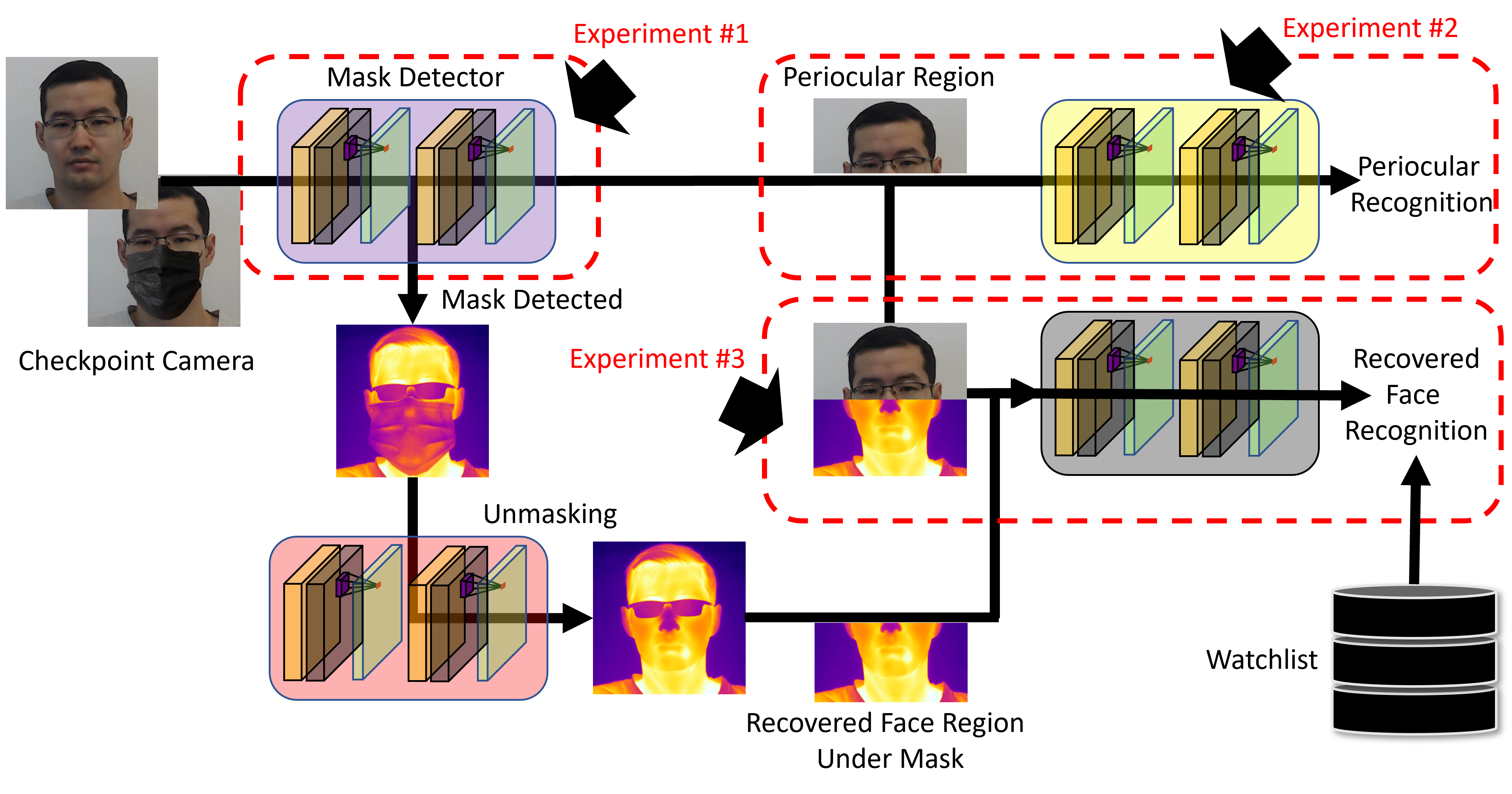}
			\caption{Adjustment of the biometric recognition mode of the post-pandemic checkpoint: the cameras in both visible and infrared bands are used. The 'recovered' face image features (transformed to a template) are compared with the face image template stored in e-ID, as well as the watchlist.
			}\label{fig:Experiment_Protocol}
		\end{center}
	\end{figure*}

	We conducted the following experiments in order to achieve the above formulated goals (Fig. \ref{fig:Experiment_Protocol}):
	
	\begin{enumerate}
		\item [] \hspace{-8mm}\emph{Experiment I:} Mouth and nose cover detection (mask detector).
		\item [] \hspace{-8mm}\emph{Experiment II:} Face verification (1:1 comparison) using a periocular region in order to estimate the face recognition performance degradation;
		\item [] \hspace{-8mm}\emph{Experiment III:} Face identification (1:N comparison) using visual, thermal, and thermal+visual hybrid image to examine whether capitalizing on different spectral domains can mitigate performance degradation.
	\end{enumerate}
	
	The main outcome of these experiments is the generation of a recovered face image using the information from both the visual and thermal domains. We assume that images of an individual (a traveler) are taken at a checkpoint using both visible spectrum and thermal camera. These images of the subject are processed in order to determine whether or not the subject is wearing a mask; this is performed using the mask detection approach (Experiment 1). If the subject is not wearing a mask, a regular visual face verification is applied to check whether the subject is on the watchlist (Experiment 2). If the face mask is detected, a thermal image of the same subject is used. The thermal face image is processed such that the masked portion of the face is recovered via the use of image translation techniques such as generative adversarial networks. This process is performed in the thermal domain; we determined experimentally that the thermal domain reveals more detail on the masked face as opposed to the visual domain. After the subject is 'unmasked' via such processing, we extract only the lower face region and concatenate it with the upper visual face region to generate a complete face image. This image can then be compared to the images in a pre-existing legacy watchlist (Experiment 3).
	
	\subsection{Datasets}

	In this paper, we perform three main experiments: mask detection, periocular face verification, and thermal+visual face identification. Since each task is different, the datasets required to train and test each experiment will also be different. We used two main datasets, Flickr-Faces-HQ (FFHQ) \cite{karras2019style} and SpeakingFaces dataset \cite{[Abdrakhmanova-2021]}. Using each of these datasets, the authors of the published works, Quieroz et al. \cite{[Queiroz-2021]} and Cabani et al. \cite{cabani2021maskedFace} synthetically added a mask to each image, thus creating the Thermal-Mask and MaskFace-Net dataset, respectfully.

\begin{figure}[!h]
		\begin{center}
			\begin{tabular}{cccc}
				\includegraphics[width=0.1\textwidth]{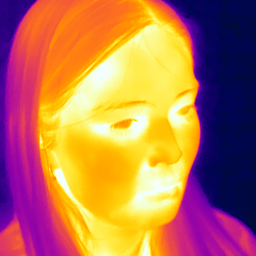}&
				\includegraphics[width=0.1\textwidth]{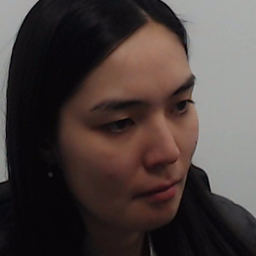}&
				\includegraphics[width=0.1\textwidth]{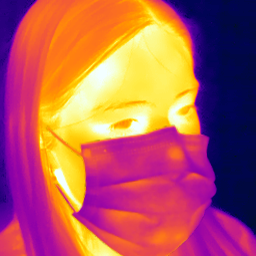}&
				\includegraphics[width=0.1\textwidth]{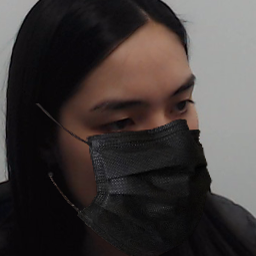}\\
				\multicolumn{2}{c}{a) Original images} &\multicolumn{2}{c}{b) Synthetic masks }\\
			\end{tabular}
			\caption{Example images of SpeakingFaces and Thermal-Mask: (a) original images from the SpeakingFaces dataset and (b) these images with synthetic masks from Thermal-Mask.}
			\label{fig:datasetth}
		\end{center}
	\end{figure}

	SpeakingFaces~ \cite{[Abdrakhmanova-2021]} and Thermal-Mask~\cite{[Queiroz-2021]} were used in Experiment 1 and 3 for face mask detection in both visible and thermal spectra. SpeakingFaces is a large-scale multimodal dataset that combines thermal, visual, and audio data streams. It includes data from 142 subjects, with a gender balance of 68 female and 74 male participants, with ages ranging from 20 to 65 years with an average of 31 years. With approximately 4.6 million images collected in both the visible and thermal spectra, each of the 142 subjects has nine different head positions and each position with 900 frames acquired in 2 trials. The Thermal-Mask dataset is a synthetically created dataset based on the SpeakingFaces. With 153,360 images, It contains both masked and unmasked subjects. Fig.~\ref{fig:datasetth} illustrates an unmasked face image (thermal + visual) and a synthetically masked face image (thermal + visual) obtained by the created algorithm, described in~\cite{[Queiroz-2021]}.

\begin{figure}[!h]
		\begin{center}
			\begin{tabular}{cccc}
				\includegraphics[width=0.1\textwidth]{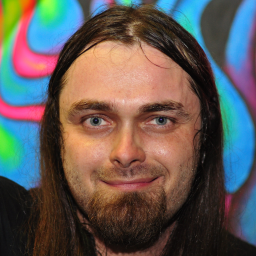}&
				\includegraphics[width=0.1\textwidth]{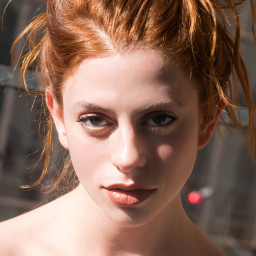}&
				\includegraphics[width=0.1\textwidth]{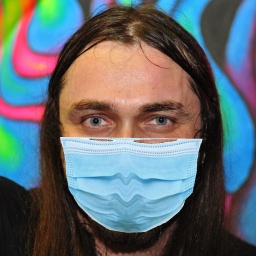}&
				\includegraphics[width=0.1\textwidth]{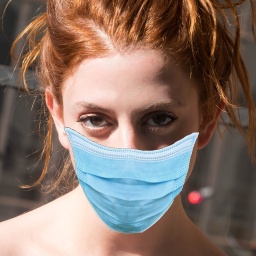}\\
				\multicolumn{2}{c}{a) Original images} &\multicolumn{2}{c}{b) Synthetic masks }\\
			\end{tabular}
			\caption{Example images of FFHQ and MaskedFace-Net: (a) original images from the FFHQ dataset and (b) these images with synthetic masks from MaskedFace-Net.}
			\label{fig:dataset}
		\end{center}
	\end{figure}

	FFHQ \cite{karras2019style} and MaskedFace-Net \cite{cabani2021maskedFace} were used in Experiment 2 for illustrating the impact of wearing masks on facial verification. FFHQ is a dataset containing 70,000 high-quality images crawled from Flickr. Images are of different subjects across varying ages, ethnicity, background, and wearing different accessories. The resolution of each image is 1024$\times$1024. The MaskedFace-Net dataset is a synthetically created dataset using FFHQ images as a base. Each image from the MaskedFace-Net dataset contains subjects either wearing a mask correctly or incorrectly. Figure \ref{fig:dataset} illustrates two images of the original subject and two images with the masks artificially 'placed over'.
	
	\subsection{Metrics}
	
	In this paper, we present three unique experiments to illustrate the checkpoint biometric recognition process. Depending on the task, different performance measures are used to report the final results. In addition, each deep learning algorithm requires a unique loss metric for it to be optimally trained.$\blacksquare$
	
	To detect masked and unmasked faces, we used the Cascade R-CNN model. As an object detector, the Cascade R-CNN applies two loss functions: the binary cross-entropy loss for classification and the smooth-L1 loss for bounding box regression. Binary cross-entropy loss is used in two-class classification tasks, such as classifying between masked and unmasked faces. It measures the error between the ground truth and the predicted value, defined by the equation below and described in~\cite{[Murphy-2012]}:
	\begin{equation}\label{eq.cross-entropy}
		\texttt{Loss\_cls}=-Y\times (log(Y_{pred}))-(1-Y)\times (log(1-Y_{pred})) 
	\end{equation}
	
	where $Y$ is the binary label, $Y_{pred}$ is the predicted value and $\log$ is the natural logarithmic function.
	
	Smooth-L1 loss is used for box regression in the Cascade R-CNN. This loss is less sensitive to outliers than most regression losses. It is defined by the equations below as described in~\cite{[Girshick-2015]}
	\begin{equation}\label{eq.smooth-l1_p1}
		\texttt{Loss\_box}=\sum_{i\in\{x,y,w,h\}}\texttt{smooth\_L1}(t_{i}^u - v_i)
	\end{equation}

	\begin{equation}\label{eq.smooth-l1_p2}
		\texttt{smooth\_L1(x)}=
		\begin{cases}
			0.5x^2,& \text{if } |x|\leq 1\\
			|x|-0.5,& \text{otherwise}
		\end{cases}
	\end{equation}
	
	where $u\in0,1,...K$ is the true class label, $t^u = (t_{x}^u, t_{y}^k, t_{w}^k, t_{h}^k)$ is the predicted bounding box correction, $v = (v_{x}, v_{y}, v_{w}, v_{h})$ is the true bounding box. 
	
	We use a multi-task \texttt{Loss} on each labeled region of interest to jointly train for classification and bounding-box regression:
	\begin{equation}\label{eq.smooth-l1_final}
		\texttt{Loss}=\texttt{Loss\_cls} + \texttt{Loss\_box}
	\end{equation}
	
	As opposed to using typical cross-entropy loss for binary decision, we employed the contrastive loss, which is shown in many works to perform well for Siamese networks. Contrastive loss is a better metric for comparison tasks as it focuses on learning from the distance metric as opposed to cross-entropy loss which focuses on classification error. The contrastive loss is defined as follows {\bf \cite{hadsell2006}}:
	\begin{equation}\label{eq.contrastive}
		\texttt{Loss}=Y\times D^2+(1-Y)
		\times \max(\texttt{Margin}-D,0)^2
	\end{equation}
	where $Y$ is the binary label, $D$ is the Euclidean distance between the two feature vectors, $\max$ is the maximum function choosing between the two provided values, and \texttt{Margin} represents the radius of influence.
	
	For object/mask detection, we used the Intersection over Union (IoU) measure to assess how well the object location prediction is. It describes the extension area of the overlap of the ground truth bounding box and the predicted bounding box.
	\begin{equation} \label{eq:iou}
		\texttt{IoU} = \frac{\texttt{Area\ of\ overlap}}{\texttt{Area\ of\ union}}
	\end{equation}
	
	The IoU is a value within the range of 0 to 1, with 1 being a perfect match between the ground truth and the predicted bounding boxes. Under this context, the true positives $TP$ will be the predicted bounding boxes whose IoU is above a chosen threshold (usually 0.5), and the false positive $FP$ occurs once the IoU is below this threshold.
	
	Based on the IoU, when ground truth is present in the image, and the model fails to detect the object, we classify it as False Negative (FN). True Negative (TN) accounts for every part of the image where we did not predict an object; however, this metrics is not useful for object detection and will be ignored.
	
	We used the mean Average Precision ($mAP$) to evaluate how well the Cascade R-CNN performs on the mask detection task.
	
	Precision represents a fraction of the relevant instances (true positives) and the total number of detected instances:
	
	\begin{equation} \label{eq:pre}
		\texttt{Precision} = \frac{TP}{TP + FP}
	\end{equation}
	
	Recall stands for a fraction of the true positive cases out of the number of ground truth cases (both true positives and false negatives):
	
	\begin{equation} \label{eq:rec}
		\texttt{Recall} = \frac{TP}{TP + FN}
	\end{equation}
	
	To assess the entire model performance, we applied the mean Average Precision ($mAP$). We first calculate the average precision $AP$ as the area under the curve (AUC) of the precision-recall curve for each category. It computes the average value of precision over the interval from $\texttt{Recall} = 0$ to $\texttt{Recall} = 1$ of $\texttt{Precision} = p(r)$. Next, we calculate the average for each category which is represented by the equation below, given the number of categories (classes) $N$:
	
	\begin{equation} \label{eq4}
		\begin{split}
			mAP & = \frac{1}{N}\sum_{i=1}^{N}	\int_{0}^{1} p_i(r_i)dr \\
		\end{split}
	\end{equation}
	
	The performance in terms of accuracy for face identification and verification is defined as follows \cite{fawcett2006}:
	\begin{equation}\label{eq:acc}
		\texttt{Accuracy} = \frac{TP+TN}{TP+TN+FP+FN}
	\end{equation}
	where $TP$ is the number of true positives, i.e. the number of matching image pairs correctly identified by the network; $TN$ is the number of true negatives, i.e. the number of non-matching image pairs correctly rejected by the network; $FP$ is the number of false positives, i.e. the number of non-matching image pairs incorrectly accepted by the network; and $FN$ is the number of false negatives, i.e. the number of matching image pairs incorrectly rejected by the network.
	
	\subsection{Experiment 1: Mouth and nose cover detection }

	In this experiment, we focused on detecting faces in the visual and infrared (thermal) spectra and classifying these faces between masked and unmasked. We used the Thermal-Mask dataset~\cite{[Queiroz-2021]} with unmasked and synthetically masked face images in the thermal spectrum that was created based on the SpeakingFaces dataset~\cite{[Abdrakhmanova-2021]}. We applied the same approach to the visual spectrum images of the SpeakingFaces, and for this study, we consider the set of all images (visual + thermal) as the Thermal-Mask dataset. 
	
	Given 142 subjects, each recorded using 9 different head positions, we selected 42,460 masked face images (80 subjects) and 33,448 unmasked face images (62 subjects) for each spectrum as described below, with the total number of images being 151,816:
	
	\setlength{\columnsep}{-70pt}
	\begin{multicols}{2}
		\begin{itemize}
			
			\item visual unmasked
			\item visual masked
			\item thermal unmasked
			\item thermal masked
			\item[] 33,448 images
			\item[] 42,460 images
			\item[] 33,448 images
			\item[] 42,460 images
			
		\end{itemize}
	\end{multicols}
	
	With the deep learning approach, we randomly selected 70\% for training (100 subjects), 20\% (28 subjects) for validation and 10\% (14 subjects) for testing, among the 142 subjects. Note that these percentages may not be precise since the subjects do not have the same amount of images after the data cleaning. Table~\ref{table:IV} summarizes the total number of samples in the final subset of the Thermal-Mask dataset, and this amount is the same for both visual and thermal versions.
	
	\begin{table}[h]
		\caption{Number of samples in each of the train, validation, and test data splits for the Thermal-Mask Dataset (for one spectrum).}
		\centering
		\begin{tabular}{c|c|c|c}
			\hline
			\textbf{Set} & \textbf{Unmasked Faces} & \textbf{Masked Faces} & \textbf{Subjects}\\[0.1em] 
			\hline
			{Train} & 23,188 & 29,842 & 100\\ 
			{Validation} & 5,940 & 8,905 & 28 \\ 
			{Test} & 4,320 & 3,713 & 14\\ 
			{Total} & 33,448 & 42,460 & 142 \\ 
			\hline
		\end{tabular}
		
		\label{table:IV}
	\end{table}

\begin{figure*}[!h]
		\centering
		\includegraphics[scale=0.9]{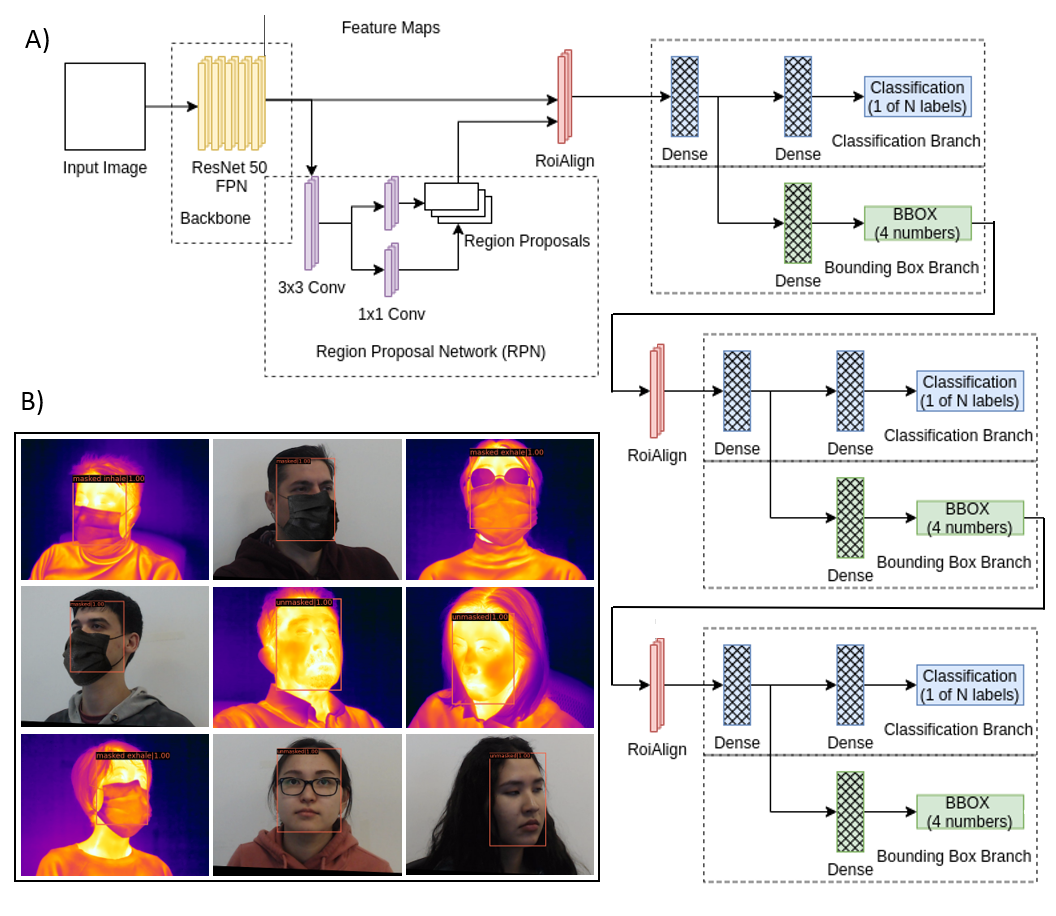}
		\caption{A) An architecture of the Cascade R-CNN Model, and B) The outputs of the Cascade R-CNN applied to the test set.} 
		\label{fig:cascade-rcnn}
	\end{figure*}

	To detect the masked and unmasked faces in thermal and visual spectra, we applied the state-of-the-art Cascade R-CNN~\cite{[cai-2019]} model separately for each spectrum. We chose a two-stage model rather than a one-stage model, focusing on accuracy over processing speed. The Cascade R-CNN is an object detector and works as a multi-stage extension of the Faster R-CNN architecture. It is composed of a sequence of detectors trained with increasing IoU thresholds. Those detectors are trained sequentially and use the output of one as the training set for the next, as seen in Fig.~\ref{fig:cascade-rcnn}A. Fig.~\ref{fig:cascade-rcnn}B illustrates some of the results of this model applied to the test set of the Thermal-Mask dataset. 
	
	To assess the effectiveness of the Cascade R-CNN model, we applied the mean Average Precision (mAP) metric, which jointly assesses the localization of the faces through bounding boxes position and the classification between masked and unmasked.~\cite{[zhu-2004]} 
	
	Table~\ref{tab:performance_3} reports the performance of the Cascade R-CNN for the mask detection task with images in visual and infrared (thermal) spectra. It illustrates the results of the applied model with four different backbones: ResNet-50, ResNet-101~\cite{[he-2016]}, ResNeXt-101-32x4d and ResNeXt-101-64x4d~\cite{[xie-2017]}. For each backbone, was applied the Feature Pyramid Network (FPN), to overcome the low resolution of the feature maps in upper layers.

	\begin{table}[!htb]
		\centering
		\caption{Cascade R-CNN}
		\label{tab:performance_3}
			\begin{tabular}{@{}c|cccc@{}} 
				\hline
				&\textbf{Backbone} & \textbf{mAP} & \textbf{mAP\textsubscript{50}} & \textbf{mAP\textsubscript{75}} \\ \hline 
				\multirow{4}{*}{\rotatebox[origin=c]{90}{Thermal}}	
				& {ResNet-50-FPN}	 & 0.873 & 0.997 & 0.986 \\
				& {ResNet-101-FPN}	 & 0.873 & 0.997 & 0.990 \\
				& {ResNet-101-32x4d-FPN} & 0.877 & 0.997 & 0.989 \\
				& {ResNeXt-101-64x4d-FPN} & \cellcolor{green}0.879 & \cellcolor{green}{0.997} & \cellcolor{green}0.990 \\
				\hline
				\multirow{4}{*}{\rotatebox[origin=c]{90}{Visual}}	
				& {ResNet-50-FPN}	 & 0.945 & 0.995 & 0.994 \\
				& {ResNet-101-FPN}	 & 0.947 & 0.995 & 0.995 \\
				& {ResNeXt-101-32x4d-FPN} & \cellcolor{green}{0.951} & \cellcolor{green}{0.995} & \cellcolor{green}{0.995} \\
				& {ResNeXt-101-64x4d-FPN} & 0.945 & 0.995 & 0.995 \\
				
			\end{tabular}
	\end{table}

	The $mAP$ column in Table~\ref{tab:performance_3} indicates the results of the $mAP$ calculated over different IoU thresholds (0.5:0.05:0.95), with the averages over all classes and also over the IoU thresholds. The subsequent columns shows the $mAP$ applied to IoU equals to 0.5 (mAP\textsubscript{50}) and 0.75 (mAP\textsubscript{75}).
	
	The ResNeXt~\cite{[xie-2017]} CNN contains particular attributes, such as parallel paths, which presents better performance compared to ResNet~\cite{[he-2016]} with the same complexity (number of floating point operations - FLOPS). It happens because the ResNeXt topology shares the same hyperparameters (width and filter sizes) between the parallel modules, reducing the total number of hyperparameters. Based on this observation, we can further improve the performance by increasing the cardinality (number of parallel paths) rather them increasing the number of layers. In the thermal spectrum, better results were obtained with ResNext with $\texttt{cardinality} = 64$ and a $\texttt{bottleneck}\_\texttt{width} = 4d$ in all metrics. However, in the visual spectrum, we observed a slight difference that made the model with lower $\texttt{cardinality} = 32$ better in the $mAP$ with multiple IoUs. The authors of the ResNeXt backbone mentioned in the original article that in some cases, increased cardinality will begin to show a saturation of accuracy for more complex datasets. Since the visible spectrum images are relatively more complex than the infrared spectrum (thermal) images, we believe that $\texttt{cardinality} = 32$ got better results than $\texttt{cardinality} = 64$.

	The above leads to the following {conclusion:} With the Cascade R-CNN model, we can locate and classify faces with or without masks at a relatively high performance (0.879 $mAP$). These results provide sufficient precision and can be given to the face identification/verification module for further processing.

	\subsection{Experiment 2: Periocular recognition and performance degradation}
	
	In this experiment, we explore the influence of wearing masks on facial verification (1:1 comparison) using the FFHQ \cite{karras2019style} and MaskedFace-Net \cite{cabani2021maskedFace} datasets. For this experiment, we designed 3 training cases and 3 testing cases, resulting in 9 performance measures. We have the following image pairs for testing/training:
	\begin{itemize}
		\item (a) image 1: no mask, image 2: no mask
		\item (b) image 1: mask, image 2: mask
		\item (c) image 1: no mask, image 2: mask
	\end{itemize}
	where no mask (FFHQ) represents an image containing a subject that is not wearing a mask and mask (MaskedFace-Net) represents an image containing a subject wearing a mask. 
	
	For performance evaluation, we used a 5-fold cross-validation method where all the samples were divided into 5 partitions. For each fold, 4 partitions are used for training while the remaining partition is used for validation. The results are then averaged across the 5 folds. 
	
\begin{figure}[!h]
		\begin{center}
			\includegraphics[width=0.5\textwidth]{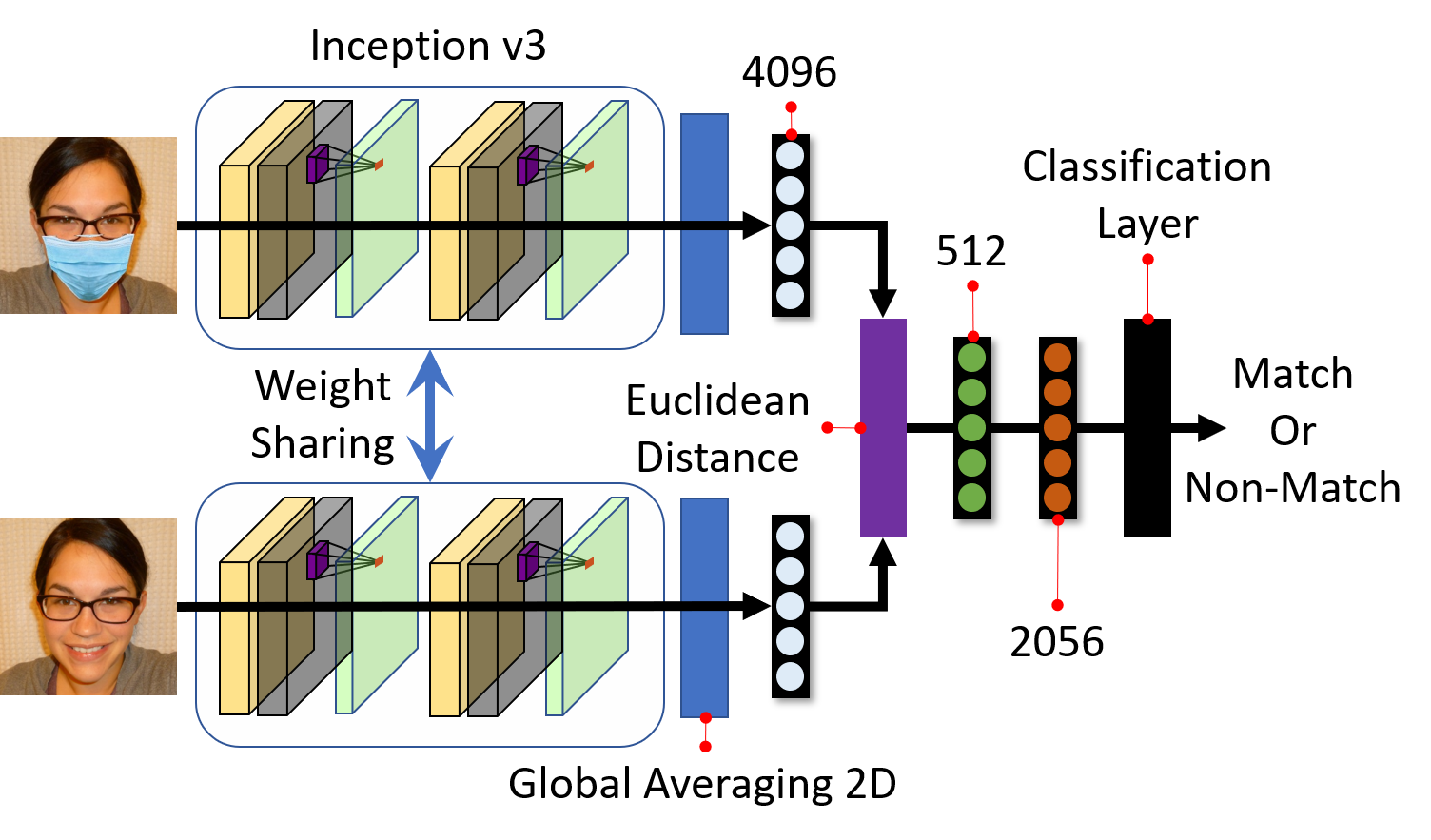}
			\caption{A Siamese Network for face verification. Inception v3 is used as the backbone network to extract image features. The features are then compared via Euclidean distance to measure the similarity between the 2 features. If the distance is small then the network determines a match between the 2 images.} \label{fig:siamese}
		\end{center}
	\end{figure}

	Due to the limited amount of images per subject, we choose to use a Siamese network \cite{dey2017signet} to perform facial verification using one-shot learning. Fig. \ref{fig:siamese} illustrates the overall architecture used in this paper for facial matching given two images. The base idea of a Siamese network is to extract features from two images then compare the features from each image to determine whether they are of the same person. The comparison task can be performed using a distance metric such as the Euclidean distance. Ideally, when two images are of the same subject, the extracted features from each image should be similar and therefore the Euclidean distance between the two should be near-zero, the opposite should be true when comparing different subjects. Note that since the Siamese network compares 2 images, the type of image for image 1 and image 2 does not matter, ie. image 1 and image 2 can be reversed.
	
	The Siamese network is trained using the Adam optimizer for 10 epochs, further details regarding the training process and hyper-parameter turning can be found in Appendix \ref{sec:app_2}.
	
	Table \ref{tab:performance_1} illustrates the accuracy (Equation \ref{eq:acc}) performance of facial verification given the various training and testing scenarios. It can be seen that when the training and testing set is the same, the performance is near perfect ($>98\%$), while different sets result in much-degraded performance ($<64\%$). This observation is likely due to the network model being overfitted to specific elements discovered between the images but is obscured when a mask is worn. This results in a degraded network that is no longer capable of matching these image pairs. This degradation indicates that by default the network train with uncovered faces is incapable of performing facial verification on individuals wearing masks. Note that for train case (c), all test results report fairly high accuracy, indicating that when possible, it is beneficial to train on both mask/unmasked subjects; however, the case of training on mask/unmasked subjects may not always be possible.
	\begin{table}[!htb]
		\centering
		\caption{Verification Accuracy using full-face images}
		\label{tab:performance_1}
		\begin{footnotesize}
			\begin{tabular}{@{}cc|ccc@{}} 
				&&\multicolumn{3}{c}{Test}\\ 
				&& (a) & (b)	&	(c) \\
				\hline 
				\multirow{3}{*}{\rotatebox[origin=c]{90}{Train}}	&	(a)	&	0.9978	$\pm$	0.0025	&	0.9951	$\pm$	0.0064	&	0.6300	$\pm$	0.0675	\\
				&	(b)	&	0.9949	$\pm$	0.0054	&	0.9974	$\pm$	0.0017	&	0.6339	$\pm$	0.1701	\\
				&	(c)	&	0.9880	$\pm$	0.0060	&	0.9891	$\pm$	0.0067	&	0.9842	$\pm$	0.0047	\\
			\end{tabular}
		\end{footnotesize}
	\end{table}
	
	Next, we examine the performance of the same network given images with only the periocular region shown. The details of creating the modified (periocular) images are provided in Appendix \ref{sec:app_2}. Table \ref{tab:performance_2} reports the performance of the Siamese network when using the periocular images for training and testing.
	\begin{table}[!htb]
		\centering
		\caption{Verification Accuracy using modified (periocular) images}
		\label{tab:performance_2}
		\begin{footnotesize}
			\begin{tabular}{@{}cc|ccc@{}} 
				&&\multicolumn{3}{c}{Test}\\ 
				&& (a) & (b)	&	(c) \\
				\hline 
				\multirow{3}{*}{\rotatebox[origin=c]{90}{Train}}	&	(a)	&	0.9990	$\pm$	0.0010	&	0.9989	$\pm$	0.0010	&	0.9934	$\pm$	0.0030	\\
				&	(b)	&	0.9932	$\pm$	0.0113	&	0.9932	$\pm$	0.0114	&	0.9881	$\pm$	0.0108	\\
				&	(c)	&	0.9975	$\pm$	0.0023	&	0.9974	$\pm$	0.0023	&	0.9955	$\pm$	0.0031	\\
			\end{tabular}
		\end{footnotesize}
	\end{table}
	The results presented in Tables \ref{tab:performance_1} and \ref{tab:performance_2} show that degradation of the different training/testing sets is greatly reduced when using periocular images. Directly comparing row (a) and column (c) from Table \ref{tab:performance_1} and \ref{tab:performance_2}, we get 0.6300 and 0.9934, respectively. Taking the difference, we get a verification accuracy difference of \fbox{0.3634}, illustrating the degradation from wearing a mask can be greatly compensated by using the periocular region for verification.
	
	To summarize, the conducted experiments illustrate the current limitation of one-shot face verification models, specifically when subjects are wearing masks. The accuracy becomes 36.78\% (99.78\% in recognizing faces of the subjects not wearing a mask which degrades by 63.00\% when the subjects are wearing a mask). A possible remedy as confirmed by the experiments indicates that performing face verification with emphasis on the periocular region can greatly alleviate this problem. The difference between periocular biometrics and masked individuals, we observe an accuracy difference of 36.34\% (99.34\% with periocular which is 63.00\% less for the faces of subjects wearing masks). This shows that by placing emphasis on the periocular region, we are able to reduce the accuracy degradation due to masks by a huge margin without the need of a specific ``masked face'' dataset.
	
	\subsection{Experiment 3: Compensation of periocular degradation}
	
	In this experiment, we are interested in examining the performance of the recognition algorithm to work on hybrid visual and thermal face images. A hybrid image consists of two portions: the top portion that includes the forehead and the periocular region of the face taken in the visual spectrum, and the bottom portion of the image consisting of the mouth and neck region of the unmasked individual generated in the thermal spectrum. By combining the top and the bottom portion of the image, we attempt to recover a complete face image to be used for matching the database face data.
	
\begin{figure}[!h]
		\begin{center}		
			\begin{tabular}{cccc}
				\includegraphics[width=0.1\textwidth]{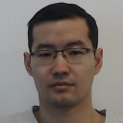}&
				\includegraphics[width=0.1\textwidth]{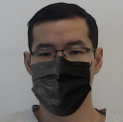}&
				\includegraphics[width=0.1\textwidth]{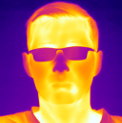}&
				\includegraphics[width=0.1\textwidth]{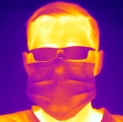}\\
				(a)	&	(b)	&	(c)		&	(d)	\\
			\end{tabular}
			\caption{Sample face images: (a) unmasked visual, (b) masked visual, (c) unmasked thermal, and (d) masked thermal image from the Thermal-Mask and SpeakingFace datasets.}\label{fig:list_image}
		\end{center}
	\end{figure}

	In Experiment 3, we focus on facial identification (1:N comparison) as opposed to Experiment 2 which focused on facial verification (1:1 comparison). As such, we choose to use the Thermal-Mask \cite{{[Queiroz-2021]}} and SpeakingFace \cite{[Abdrakhmanova-2021]} datasets as they contain facial images of subjects taken in both visual and thermal domains. For this experiment, we used five types of images to illustrate the impact of using thermal images of masked faces on the subject recognition performance. The types of images we used include: visual face, masked visual face, thermal face, masked thermal face, and recovered face image (Fig. \ref{fig:list_image}). 
	
\begin{figure}[!h]
		\begin{center}
			\includegraphics[width=0.5\textwidth]{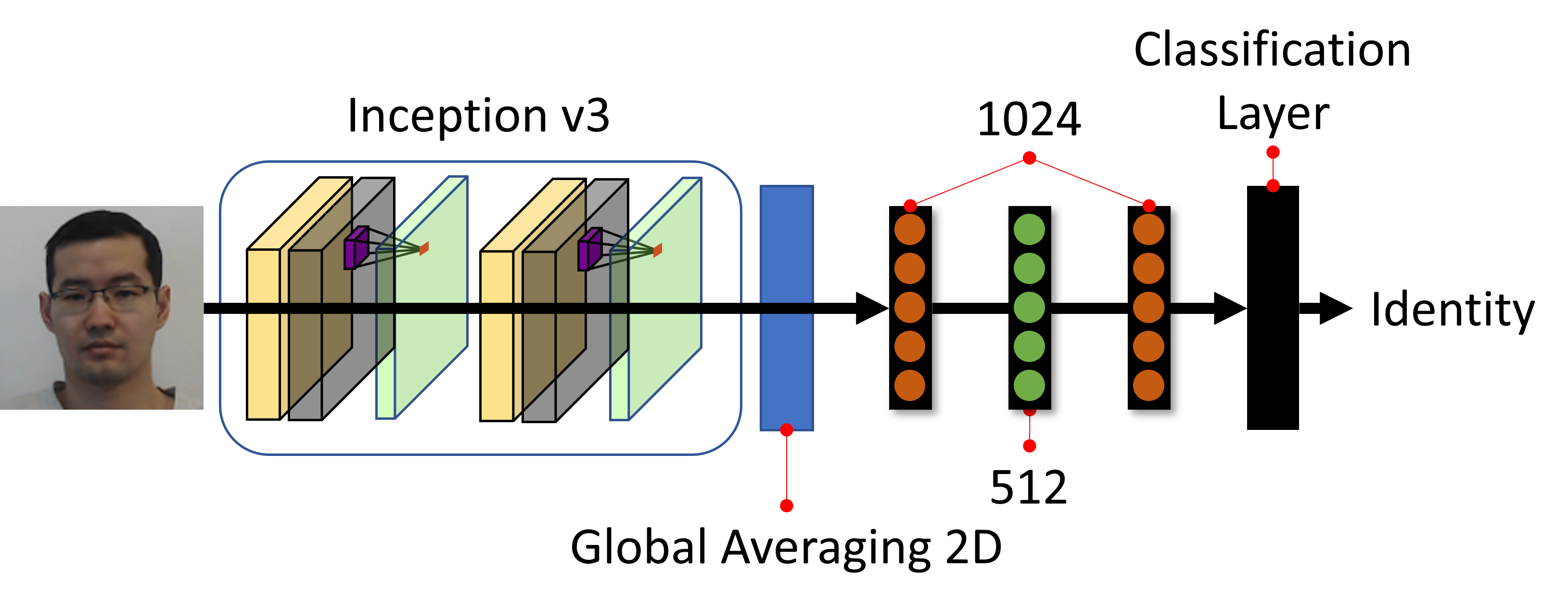}
			\caption{A CNN for face identification: Inception v3 is used as the backbone network to extract image features; the features are then processed to determine the identity of the subject.}\label{fig:cnn}
		\end{center}
	\end{figure}

	For this experiment, a typical Convolutional Neural Network (CNN) based on Inception v3 for feature extraction is used. This CNN (Fig. \ref{fig:cnn}) is trained using the Adam optimizer via the categorical cross-entropy loss function. A detailed description of the CNN architecture is provided in Appendix \ref{sec:app_3}. The performance evaluation is based on the identification accuracy (Equation \ref{eq:acc}) across 5-fold cross-validation.
	
	Table \ref{tab:performance_4} reports the identification accuracy (Equation \ref{eq:acc}) using an assorted mask and unmasked faces taken in thermal, visual, and thermal+visual domains. The column represents the image type used for testing, and the row represents the image type used for training. For example, row-1 and column-3 refer to training the CNN using unmasked visual face image and testing with unmasked thermal face image. In this experiment, we see a performance degradation of $0.9982-0.4189=$\fbox{0.5793} (taken from row-1, column-1 and row-1, column-2 in Table \ref{table:IV}) in the visual domain when a mask is worn. Similarly, a loss of $0.9899-0.3055=$\fbox{0.6844} (taken from row-3, column-3 and row-3, column-4 in Table \ref{table:IV}) is observed in the thermal domain.
	
	\begin{table*}[!htb]
		\centering
		\caption{Identification accuracy using images from the Thermal-Mask and SpeakingFace datasets}
		\label{tab:performance_4}
		\begin{footnotesize}
			\begin{tabular}{@{}cc|ccccc@{}} 
				&&\multicolumn{5}{c}{Test}\\ 
				&& \includegraphics[width=0.05\textwidth]{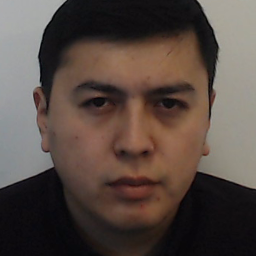} & \includegraphics[width=0.05\textwidth]{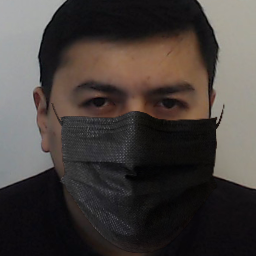}	&	\includegraphics[width=0.05\textwidth]{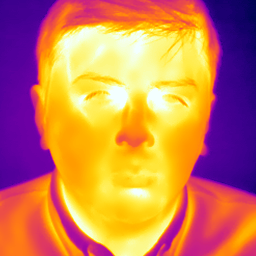} &	\includegraphics[width=0.05\textwidth]{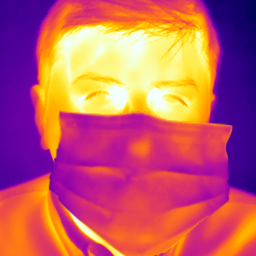} &	\includegraphics[width=0.05\textwidth]{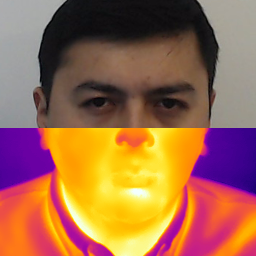}\\
				\hline 
				\multirow{10}{*}{\rotatebox[origin=c]{90}{Train}}	&	\includegraphics[width=0.05\textwidth]{fig/unmasked_visual.png}	&	0.9982	$\pm$	0.0009	&	0.4189	$\pm$	0.1962	&	0.0111	$\pm$	0.0037	&	0.0118	$\pm$	0.0013	&	0.2220	$\pm$	0.2199	\\
				&	\includegraphics[width=0.05\textwidth]{fig/masked_visual.png}	&	0.8334	$\pm$	0.1472	&	0.9899	$\pm$	0.0079	&	0.0124	$\pm$	0.0047	&	0.0121	$\pm$	0.0047	&	0.2123	$\pm$	0.3280	\\
				&	\includegraphics[width=0.05\textwidth]{fig/unmasked_thermal.png}	&	0.1079	$\pm$	0.0291	&	0.0175	$\pm$	0.0077	&	0.9899	$\pm$	0.0093	&	0.3035	$\pm$	0.1328	&	0.1079	$\pm$	0.0291	\\
				&	\includegraphics[width=0.05\textwidth]{fig/masked_thermal.png}	&	0.0191	$\pm$	0.0049	&	0.0172	$\pm$	0.0049	&	0.8421	$\pm$	0.1234	&	0.9937	$\pm$	0.0033	&	0.0212	$\pm$	0.0073	\\
				&	\includegraphics[width=0.05\textwidth]{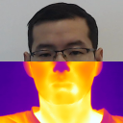}	&	0.6362	$\pm$	0.1057	&	0.4704	$\pm$	0.0503	&	0.1111	$\pm$	0.0505	&	0.0300	$\pm$	0.0177	&	0.9803	$\pm$	0.0162	\\
			\end{tabular}
		\end{footnotesize}
	\end{table*}

	There are a few observations shown in Table \ref{tab:performance_4}. 
	
	\begin{enumerate}
		\item Diagonal of the matrix shows the highest performance. This exceptional accuracy is most likely because both the training and testing sets are from the same cohort (same image domain and both are wearing/not-wearing masks). 
		\item Accuracy between images from the same image domain (visual or thermal) is much higher than cross-domain (visual vs. thermal). 
		\item Masked images are used for training and are tested on unmasked images, its performance is magnitudes higher than the reverse.
	\end{enumerate}
	
	The numbers shown in row-1, column-2 and row-5, column-2 illustrate the scenario where the models are trained on a pre-existing dataset (watchlist), and its accuracy only reaches 41.89\% when the same individual wears a mask. This accuracy can be improved to 47.04\% if the bottom half mouth region (hidden under the mask in the visual domain) is replaced with a thermal region of the mouth (synthetically unmasked). Note that this compensation method via hybrid images is only applicable in the visual domain, as the periocular region in the thermal domain is often obscured by the use of clear glasses (which appear to be dark in thermal but clear in visual).

	\section{Summary, conclusion and future work}\label{sec:conclusion}
	
	Leaders in security technologies such as NEC corporation have been on the search for epidemic-conditioned solutions in biometrics, and for ways to overcome the degradation of face recognition. Recently, the NEC reported that they achieved 99.9 \% accuracy of periocular recognition compared with 98.21\% reported in our experiments. However, the NEC achieved these results through the process of QR immunity passport verification, i.e. in scenarios when additional information is available. Specifically, the NEC combined three authentication methods: associated possession (ID), token, and a biometric trait, thus reducing the problem of biometric identification to verification, i.e. reducing from 1:\texttt{Many} to 1:1 matching \cite{[Bolle-2004]}. Based on our experimental results, we deduce that it is realistic to recover the accuracy that we achieved in our experiments, by 1.79\% in order to achieve 99.9 \% as reported by NEC.

	\section*{Acknowledgments}
	
	\begin{small}
		This Project was partially supported by Natural Sciences and Engineering Research Council of Canada (NSERC) through grant ``Biometric intelligent interfaces''.
	\end{small}

	\section{Appendix}
	
	This Appendix provides more detail on the parameters of the experiments described in this paper.
	
	\subsection*{Experiment 1} \label{sec:app_1}
	\subsubsection*{Parameters}
	
	For this experiment, our model was trained for 12 epochs using a mini-batch size of 2 and the Stochastic Gradient Descent (SGD) optimizer, with the parameters learning rate of $0.002$, $\beta_1=0.9$, and $\beta_2=0.0001$. The learning rate is a tuning hyper-parameter that determines the step size at each iteration while minimizing the loss function. The $\beta_1$ stands for the momentum, which adds a fraction of the previous weight update to the current one to avoid local minima and speeds up the training time. The $\beta_2$ stands for the weight decay, a regularization technique that adds a small penalty to the loss function. As an object detector, the model applies two loss functions: the binary cross-entropy loss~\cite{[Murphy-2012]} for classification and the smooth-L1 loss~\cite{[Girshick-2015]} for the bounding box regression (localization). 
	
	\subsubsection*{Architecture}
	For this experiment, we divided our model architecture into the following parts:
	
	\begin{enumerate}
		\item{\textbf{Backbone,}} a CNN which takes as input the image and extracts the feature map.
		\item{\textbf{Neck,}} a component between the backbone and head of the architecture that performs improvements or refinements to feature maps.
		\item{\textbf{Head,}} a target object (masked/unmasked face) detector part of the network architecture.
	\end{enumerate}
	
	In our experiment, we applied the ResNet~\cite{[he-2016]} backbone and the ResNext~\cite{[xie-2017]}. We compared the ResNet-50 and Resnet-101 with 50 and 101 layers, respectively. For the ResNeXt, we considered the ResNeXt-101-32x4d, which stands for the architecture with 101 layers, 32 parallel pathways, and a bottleneck width of 4 dimensions. We also applied the ResNeXt-101-64x4d with higher cardinality of 64.
	
	A sequence of several CNN layers usually leads to an increase in the semantic value of feature maps, while the spatial dimension (resolution) decreases. To overcome the low resolution of the feature maps in the upper layers, we applied the Feature Pyramid Network (FPN)~\cite{[Lin-2017]}. It takes an image as an input and outputs the feature maps at multiple levels (different sizes) in a fully convolutional fashion, which improves the detection of small objects.
	
	The head of our architecture is the Cascade R-CNN~\cite{[cai-2019]}, which is composed of a sequence of detectors trained with increasing Intersection-of-Unions (IoU) thresholds. It is implemented with four stages: one Region Proposal Network (RPN) and three detection heads with thresholds IoU = {0.5, 0.6, 0.7}.

	\subsection*{Experiment 2} \label{sec:app_2}

	\subsubsection*{Parameters}
	For this experiment, we used the Adam optimizer to train our Siamese network. Adam \cite{kingma2017adam} is an optimization algorithm used to replace the standard stochastic gradient descent. We chose the following parameters for the training: a learning rate of $0.001$, $\beta_1=0.9$, $\beta_2=0.999$, an epoch of 10, and a batch size of 32. Learning rate is a hyper-parameter that controls how much update is applied to the model. $beta_1$ and $beta_2$ are 2 coefficients used to control the decay rate of the first and second moments, respectively. Epoch is the number of times the entire training set is used for training. Batch size is a hyper-parameter that determines how many samples the model sees before an update is triggered.
	
	\subsubsection*{Architecture}
	In this experiment, the machine-learning model is the Siamese network designed to perform 1-to-1 image comparisons corresponding to the facial verification (1:1 comparison). A unique component of this Siamese network is the ``twin'' connection between the two backbone networks. For this experiment, we used Inception v3 as the backbone network, which is a CNN proposed by Szegedy et al. \cite{Szegedy2016rethinking}. It is pre-trained on the ImageNet dataset, to extract image features. The ``twin'' connection is designed in such a way that both networks share the same exact weights and, therefore, when presented with the same image, they should yield the exact same output. Each output from these two backbone networks is passed through a global average pooling layer, and a 2096-unit fully-connected layer. The Euclidean distance is then calculated between the output of the 2 2096-unit fully-connected layers. The computed distance is then analyzed through 3 fully-connected layers with unit size 512, 2056, and 2. The resulting output of the 2-unit is a 2-class probability representing match or no-match.

	\subsubsection*{Image Rescaling}
	The images used in this experiment are from the FFHQ dataset which contains images taken in 1024$\times$1024 pixel resolution. Since for this experiment the image resolution is not a strict requirement, all images have been scaled down to 256$\times$256 pixel resolution got the purpose of conserving the memory usage. The rescaling process is done via area-based interpolation.
	
	\subsubsection*{Periocular Processing}
\begin{figure}[!h]
		\begin{center}
			\begin{tabular}{cc}
				\includegraphics[width=0.22\textwidth]{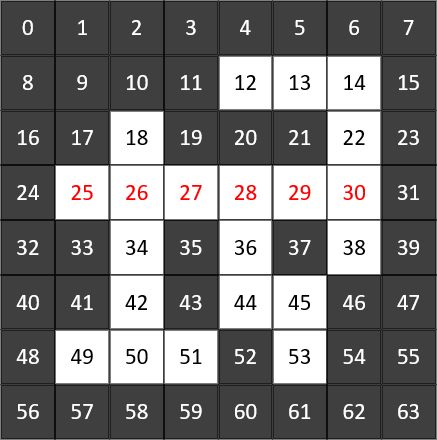}&
				\includegraphics[width=0.22\textwidth]{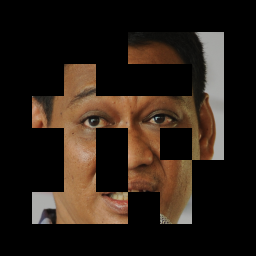}\\
				(a) & (b) \\
			\end{tabular}
			\caption{Example of a modified face image: (a) dividing the face to $8 \times 8$ regions, and (b) applying the mask to a random subject.}\label{fig:blackout}
		\end{center}
	\end{figure}

	The creation of periocular images is done via ``blacking out'' the non-essential regions of the face. The proposed modification is to divide each image into equal 8x8 regions. After the image rescaling process, each image has a pixel resolution of 256$\times$256. We grouped 32x32 pixels together, to divide the image into equal 8x8 regions and labeled them sequentially (with the top left corner being labelled 0, and the bottom right corner being labelled 63). Since the original authors centered the location of the detected face, we determine that the periocular region of the face is located between regions 25-30 (highlighted red in Fig. \ref{fig:blackout} (a)). 
	
	In this paper, we propose to use a masking procedure to perform the ``blacking out'' process. The purpose of such a masking technique is to maintain the same image resolution as the original image while randomly ``blacking out'' the non-periocular regions of the face. By removing or ``blacking out'' these regions, we are essentially steering any networks using these images to focus on the periocular regions as opposed to other features on the face (such as the nose). The mask shown in Fig. \ref{fig:blackout} contains the following properties: 
	\begin{itemize}
		\item borders of the image are always dropped (top/bottom rows and left/right columns are always dropped).
		\item periocular regions are always kept intact (regions 25 to 30).
		\item all remaining regions have a 50\% chance of getting dropped.
	\end{itemize}
	
	\subsection*{Experiment 3} \label{sec:app_3}
	
	\subsubsection*{Parameters}
	For this experiment, we used the Adam optimizer \cite{kingma2017adam} to train our CNN. Similar to Experiment 2, the following parameters were used: a learning rate of $0.001$, $\beta_1=0.9$, $\beta_2=0.999$, an epoch of 10, and a batch size of 32.
	
	\subsubsection*{Architecture}
	In this experiment, our machine-learning model is a transfer learning-based CNN such as Inception v3 \cite{Szegedy2016rethinking}. The transfer learning process involves modifying the original Inception v3 by replacing the top fully-connected layers with new fully-connected layers. The network is then fine-tuned with these new fully-connected layers in order to adapt the network to new data. For this experiment, we replaced the original layers with a global average pooling, 1024-unit fully-connected, 512-unit fully-connected, 1024-unit fully-connected, and 80-unit fully-connected. The global average pooling layer averages the features across the channel dimensions, thereby converting the feature vector from 2D to 1D. The 1024-512-1024 unit configuration is a simple multilayer perceptron shown to yield the best performance in this experiment. The last 80-units (chosen because we had 80 subjects) in a fully-connected layer represent the classification layer which outputs the identity of the subject.
	
	\subsubsection*{Image Rescaling}
	The images used in this experiment are from the SpeakingFace and Thermal-Mask dataset which contains images taken in various pixel resolutions (464$\times$348 for thermal and 768$\times$512 for visual). Similar to Experiment 2, all images are scaled to 256$\times$256 pixel resolution using area-based interpolation.
	
	\subsubsection*{Image Processing}
\begin{figure}[!h]
		\begin{center}
			\includegraphics[width=0.49\textwidth]{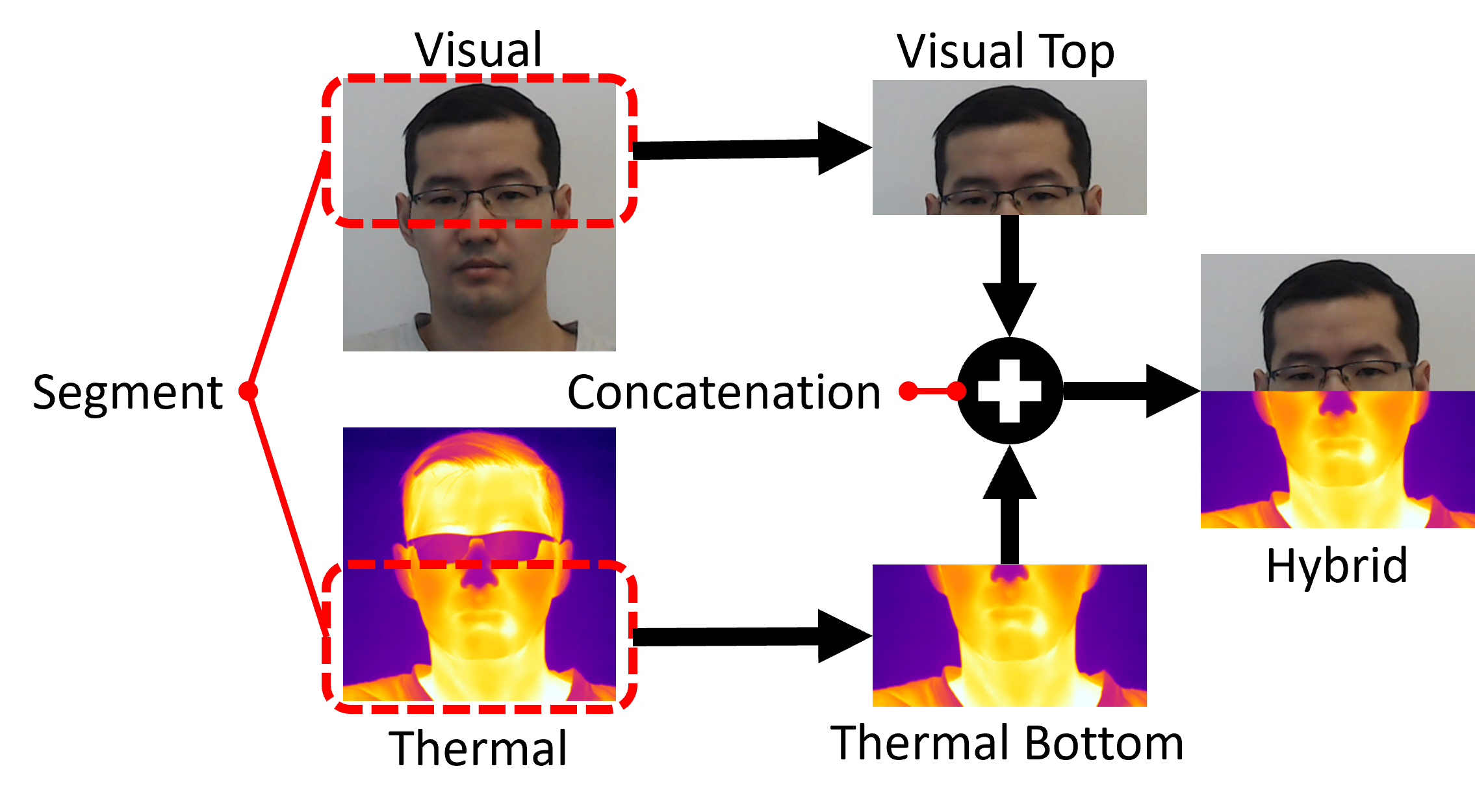}
			\caption{A process of creating a visual+thermal hybrid image: the top half is the visual image and the bottom half is the thermal image; they are concatenated together to recover a full-face image. }\label{fig:segment}
		\end{center}
	\end{figure}

	For this experiment, we deployed three image domains: thermal, visual, and thermal+visual (hybrid). Both the thermal and visual images can be processed directly, while the hybrid images require additional segmentation and concatenation. The hybrid image consists of two partial images, the top half is the visual image and the bottom half is the thermal image. Since each image is pre-aligned by the authors of the data set \cite{[Abdrakhmanova-2021]}, we were able to directly crop and use the top half of the visual image and the bottom half of the thermal image. Fig. \ref{fig:segment} illustrates the hybrid image creation process.
	
\end{document}